%% file: eccv2022submission.tex
\documentclass[runningheads]{llncs}
\usepackage{graphicx}

\usepackage{tikz}
\usepackage{comment}
\usepackage{amsmath,amssymb} 
\usepackage{color}
\usepackage{orcidlink}

\usepackage{mathtools}
\usepackage{enumitem}
\setitemize{noitemsep,topsep=0pt,parsep=0pt,partopsep=0pt}
\usepackage{multirow}
\usepackage{booktabs}
\input{macro}

\usepackage{pifont}
\usepackage{soul}

\usepackage[accsupp]{axessibility}  

\begin{document}
\pagestyle{headings}
\mainmatter
\def\ECCVSubNumber{3247}  

\title{Efficient One-stage Video Object Detection by Exploiting Temporal Consistency}

\titlerunning{Efficient One-stage Video Object Detection}
%
\author{Guanxiong Sun\inst{1,2}
\orcidlink{0000-0003-1901-9097} 
\and
Yang Hua\inst{1}
\orcidlink{0000-0001-5536-503X} 
\and
Guosheng Hu\inst{2}
\orcidlink{0000-0002-9448-9892} 
\and
Neil Robertson\inst{1}
\orcidlink{0000-0003-2461-8799}
}
\authorrunning{Guanxiong et al.}
%
\institute{EEECS/ECIT, Queen's University Belfast, UK \and
Oosto, Belfast, UK \\
\email{\{gsun02, y.hua, n.robertson\}@qub.ac.uk}, 
\email{huguosheng100@gmail.com}
}
\maketitle

\begin{abstract}
Recently, one-stage detectors have achieved competitive accuracy and faster speed compared with traditional two-stage detectors on image data. However, in the field of video object detection (VOD), most existing VOD methods are still based on two-stage detectors. Moreover, directly adapting existing VOD methods to one-stage detectors introduces unaffordable computational costs. In this paper, we first analyse the computational bottlenecks of using one-stage detectors for VOD. Based on the analysis, we present a simple yet efficient framework to address the computational bottlenecks and achieve efficient one-stage VOD by exploiting the temporal consistency in video frames. Specifically, our method consists of a location-prior network to filter out background regions and a size-prior network to skip unnecessary computations on low-level feature maps for specific frames. We test our method on various modern one-stage detectors and conduct extensive experiments on the ImageNet VID dataset. Excellent experimental results demonstrate the superior effectiveness, efficiency, and compatibility of our method. The code is available at \href{https://github.com/guanxiongsun/vfe.pytorch}{https://github.com/guanxiongsun/vfe.pytorch}.

\end{abstract}

\section{Introduction}

Recently, in the field of object detection on still image data, great attention has been paid to one-stage detectors \cite{yolox,corner,yolov3,fcos,centernet,extreme}, as they have shown many stunning advantages compared to traditional two-stage detectors \cite{rfcn,fastrcnn,maskrcnn,fasterrcnn}. For example, one-stage detectors are more efficient, straightforward, and well aligned with other fully convolutional tasks, e.g., semantic segmentation, facilitating these tasks to share ideas and tricks.
Given these advantages, many researchers work in different directions to further improve the accuracy of one-stage detectors, such as label assignment \cite{ota,atss}, feature alignment \cite{dyhead,mimicdet}, loss design \cite{general_focal_loss,focal_loss,varifocal}, and multilevel feature aggregation \cite{nasfpn,fpn,pan}. 
Until now, compared with two-stage detectors, one-stage detectors can achieve very competitive accuracy with faster run-time speed.

However, the field of video object detection (VOD) has been dominated by two-stage detectors \cite{mega,rdn,mamba,selsa,fgfa,dff} for many years and very little research has been investigated to transfer the merits of one-stage detectors from still images to videos \cite{chp}. 

This phenomenon contradicts empirical intuitions that one-stage detectors are more suitable for VOD task, which requires faster speed. To investigate the underlying reasons for this phenomenon, we conduct a comprehensive quantitative analysis (detailed in \sect{analysis}) and share the following facts:
(1) The SOTA VOD methods apply attention-based feature aggregation to achieve promising speed-and-accuracy trade-offs; 
(2) In two-stage SOTA VOD methods, the computational cost is reasonable since the attention module take a small number of proposals as inputs, e.g., 300 proposals; 
(3) Directly adapting existing VOD methods to one-stage detectors introduces unacceptably high computations, due to the drastically increased number of inputs for attention modules, for example, 13k pixels in FCOS \cite{fcos};
(4) For one-stage detectors on images or videos, the detection heads on low-level features take 80\% computations.

On the basis of the aforementioned analysis, we propose two modules to achieve an efficient one-stage video object detector by fully taking advantage of the temporal consistency of video data.
Here, 
the temporal consistency denotes the fact that objects change gradually in terms of \emph{locations} and \emph{sizes} in a sequence of consecutive frames. Inspired by this, we propose two novel modules, the location prior network (LPN) and the size prior network (SPN). Specifically, 
first, 
detected bounding boxes in the previous frame can guide the model to find regions where objects may appear in the current frame. Our LPN utilises this location prior knowledge to filter out background regions and thus reduces the computational cost.
Second, objects keep in similar \emph{sizes} within a short time. Another fact is that one-stage detectors divide objects into different levels of feature maps, and each level is responsible for detecting objects in a specific size range. 
Given the object sizes in the current frame, the proposed SPN enables our method to skip unnecessary computations on unrelated feature levels in several following frames.

In summary, our main contributions are:
\begin{itemize}

    \item  To our best knowledge, we are the first to investigate the obstacles to the development of one-stage VOD and conclude 
    two bottlenecks causing high computations: very high-dimensional input for attention modules and unnecessary computations on low-level feature levels.

    \item We propose a simple yet effective framework to achieve efficient one-stage object detection. Specifically, a location prior network (LPN) filters out background regions and a size prior network (SPN) to skip computations on unnecessary feature levels. 
    Note that our method can easily be incorporated into   various one-stage detectors.

    \item Extensive experiments are conducted on ImageNet VID datasets with various one-stage detectors, i.e., FCOS  \cite{fcos}, CenterNet  \cite{centernet} and YOLOX  \cite{yolox}. The results demonstrate that our method achieves superior speed-accuracy trade-offs and promising compatibility. 
\end{itemize}

\section{Related Work} 
\subsubsection{One-stage Detectors.} 
One-stage detectors can be classified into two categories. Firstly, key point based methods that predict pre-defined key points of objects to generate bounding boxes.
For example, CornerNet  \cite{corner} treats a bounding box as a pair of top-left corner and bottom-right corner and detects objects by grouping predicted corner pairs. ExtremeNet  \cite{extreme} predicts four extreme points (i.e., top-most, left-most, bottom-most, and right-most) and one center point to produce bounding boxes. CenterNet  \cite{centernet} detects the center point of an object bounding box. It predicts heat maps of center points and several regression values (i.e., center offset and size of the bounding box) to generate bounding boxes. In this paper, we use CenterNet as a representative of key point based one-stage detectors.

Another category of one-stage detectors is the center-based method, which regards the center pixels of an object as positives, and then predicts the distances from positives to bounding box boundaries. YOLO series \cite{yolo,yolov2,yolov3} are the most well-known center-based one-stage detector. Recently, YOLOX \cite{yolox} presents many empirical improvements to YOLO series, forming a new high-performance detector.
DenseBox  \cite{densebox} utilises a filled circle located in the center of an object and predicts four distances from each location inside the circle to the boundaries of the object bounding box. FCOS  \cite{fcos} regards all locations inside an object as positives and introduces a centerness branch to measure distances between positives to the center point of the object. The centerness branch can effectively reduce false positives in the inference stage. In this paper, we use FCOS and YOLOX as representatives of center-based one-stage detectors.

\subsubsection{Video Object Detection (VOD).}
VOD methods explore using temporal information within a video to improve the performance and the speed of single-frame detectors. Existing VOD methods can be divided into two categories: box-level methods and feature-level methods. 
Box-level methods try to refine the detection results using temporal associations of predicted bounding boxes. These methods are performed in a post-processing manner. For example, TPN \cite{tpn} and TCNN \cite{tcnn} use LSTM and tracking to model temporal associations between detected bounding boxes. SeqNMS \cite{seqnms} extends NMS to the time domain and greatly reduces false positives.
CHP  \cite{chp} proposes a heat map propagation method for CenterNet  \cite{centernet}, which makes detection results temporally smooth.
In contrast, feature-level methods are investigated to improve the accuracy of video object detection by feature enhancement and can be trained end-to-end. 
FGFA  \cite{fgfa}, MANET  \cite{manet} and THP  \cite{thp} utilise optical flow to propagate and aggregate feature maps.
SELSA  \cite{selsa}, MEGA  \cite{mega}, LRTR  \cite{lrtr} and RDN  \cite{rdn} enhance the instance features (proposals) of the current frame by reasoning the relationships between objects within a video via attention mechanisms. 


\section{Analysis of the Computational Bottlenecks in Attention-based One-stage VOD}
\label{sec:analysis}

The key reason why recent video object detection (VOD) methods \cite{mega,rdn,lrtr,selsa} achieve state-of-the-art performance is the utilisation of attention mechanisms. However, these methods are based on two-stage detectors. 
We first apply attention-based VOD methods directly to one-stage detectors and demonstrate the high computational cost in this naive adaptation.
Then, we conduct an elaborate analysis to locate the reasons for this high  cost.

\subsection{Preliminary Knowledge}
\label{subsec:preliminary}

Before we dive into a detailed analysis, we introduce some preliminary knowledge and define the necessary terms.

\begin{figure*}[t]
\begin{center}
\includegraphics[width = \textwidth]{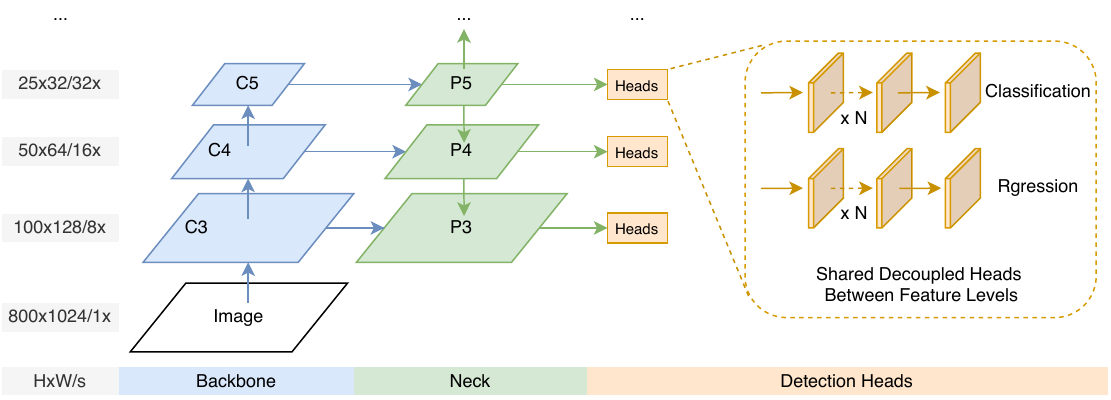}
\end{center}
\caption{General architecture of modern one-stage detectors, where H, W, and s are the height, width, and stride (down-sampling ratio) of feature maps, respectively. C3, C4 and C5 denote the output feature maps of the backbone. P3, P4, P5, etc. denote the feature levels in the neck, e.g., FPN. The decoupled detection heads, which usually contain a classification branch and a regression branch, are shared through all feature levels. \bestviewed}
\label{fig:meta}
\end{figure*}

\subsubsection{General Architecture of Modern One-stage Detectors.}
Modern one-stage detectors  \cite{yolox,focal_loss,yolov3,fcos,centernet} are designed with different modules and settings, but they share a general architecture. The general architecture can be summarised as three parts: backbone, neck, and detection head. Specifically, backbone networks extract feature maps from input images, for example, ResNet-50/101 \cite{resnet}, DarkNet-53 \cite{yolo}, and HGNet \cite{hourglass}. Then, the feature maps $\{C\}$ are forwarded into the neck module, such as FPN \cite{fpn} and PAN \cite{pan}, to conduct multi-level feature aggregation. At last, the  detection  heads are performed on all feature levels $\{P\}$ to generate detections.
A detailed general architecture is shown in \figa{meta}, where $s$ denotes the stride or down-sampling ratio of a feature level to the input image.

\subsubsection{Complexity of the Attention Module.}
\label{subsubsec:complexity_of_att}
We introduce the complexity of the attention module because it is the key of designing an efficient one-stage VOD method. 
Given a query set $\mathbf{Q}=\{q_i\} \in \mathbb{R}^{N_q \times C}$ 
and a key set $\mathbf{K}=\{k_i\} \in \mathbb{R}^{N_k \times C}$, an attention module enhances each query $q_i$ by measuring relation features as the weighted sum of all the keys in $K$. Here, $N$ and $C$ denote the number and the dimension of query or key elements, respectively. For simplicity, we use one-head attention for demonstration. Specifically, the enhanced feature of $q_i$ is:
\begin{equation}\label{equ:att}
     A(q_i, K) = q_i + \sum_j w_{i j} \cdot (W \cdot k_j),
\end{equation}
where $W$ denotes a linear transformation matrix, and $w_{ij}$ is an element in the correlation matrix computed based on the similarity of all $q$-$k$ pairs. Since the number of key elements $N_k$ is usually equal or linearly related to the number of query elements $N_q$, and the complexity of an attention module is $O({N_q}^2 \times C)$, the computational cost of the attention module is very sensitive to $N_q$.

\subsection{Naive Adaptation of Attention-based One-stage VOD}

In SOTA VOD methods, attention modules are introduced in the second stage of two-stage detectors \cite{rfcn,fasterrcnn}, where object proposals are treated as query elements. Specifically, the proposals of the current frame are considered as the query $Q$ and proposals from reference frames are regarded as the key $K$ for the attention module. Then, $Q$ is enhanced with $K$ by attention modules as \equa{att}. 
Since one-stage detectors generate proposals, a naive adaptation from SOTA methods to one-stage VOD is to conduct attention-based feature aggregation on the feature maps of one-stage detectors.
Although the idea is straightforward, problems arise because of the difference between the number of proposals and the number of pixels on feature maps.
For example, the number of proposals is quite small, e.g., 300 in FasterRCNN \cite{fasterrcnn}, while the number of pixels in one-stage detectors is usually thousands, e.g., $\sim$13K in FCOS \cite{fcos}. 
This naive adaption highly increases the computational cost of the attention module.

We design several models to quantitatively demonstrate the increased computational cost problem of the naive adaptation method. Specifically, we use SELSA \cite{selsa} as the baseline for its simplicity and promising performance. The analysis and conclusions in this subsection are suitable for other attention-based VOD methods because they have similar computation costs to SELSA.
Following SELSA, the key set $K$ is generated by randomly sampling $n_r$ reference frames in the current video.
The naive adaption is to treat all pixels on the feature maps as the query or key elements. For every time step, $Q$ and $K$ consist of pixels on feature maps of the current frame and the randomly sampled reference frames, respectively.

In the next two subsections, we analyse the experimental results and conclude two bottlenecks for the computational issue.

\begin{table}[t]
\centering
\caption{Comparisons of $N_q$ and computational costs between SELSA and directly applying attention mechanisms on one-stage detectors, FCOS, CenterNet, and YOLOX.}
\label{tab:nq}
\begin{tabular}{llcccc} 
\toprule
Method         & Input Size  & Strides $\{s\}$ & $N_q$  & GPU Mem. & FPS   \\
\midrule
SELSA          & (600, 1000) & \{16\}           & 300       &   1.8 GB  & 18.5  \\
\midrule
FCOS$^{A}$      & (600, 1000) & {\ \{8,16,32,64,128\}\ }   &  {\ 12,958\ }    & 21.9 GB  & 4.6   \\
CenterNet$^A$ & (512, 512)  & \{4\}                & 16,384 &  31.2 GB  & 4.3    \\
YoloX$^A$     & (640, 640)  & \{8, 16, 32\}        & 8,400      & 16.7 GB   & 8.9  \\
\bottomrule
\end{tabular}
\end{table}

\subsection{Bottleneck 1: Drastically Increased $N_q$}
\label{subsec:bottleneck1}

In \tab{nq}, we show the computational cost of the SELSA baseline and naive adaptions with three one-stage detectors, denoted as FCOS$^{A}$, CenterNet$^A$, and YOLOX$^A$. 
Although most SOTA VOD methods use more than 10 reference frames during inference, we only test with 2 reference frames because we suffer from GPU out-of-memory errors on Tesla V100 (32GB) GPU if more than 2 reference frames are used for naive one-stage adaption models.
For example, in SELSA \cite{selsa}, $N_q=300$ is the number of proposals of the current frame. 
Differently, in the naive one-stage adaption model, $N_q$ is related to the size of the input image.
For FCOS$^A$, we follow the protocols in SOTA VOD methods  \cite{mega,rdn,selsa} to resize the input image to a shorter side being 600 and a longer side less or equal to 1000, and thus $N_q$ is $\sim$13K. For CenterNet$^A$ \cite{centernet} and YOLOX$^A$ \cite{yolox}, following the original papers, the input images are resized to 512 $\times$ 512 and 640 $\times$ 640, and thus $N_q$ are $\sim$16.4K and 8.4K, respectively.
Compared with SELSA, $N_q$ of naive adaption models  drastically increases nearly 50 times. As analysed in \subsecs{preliminary}, the computational complexity of attention modules is quadratic to $N_q$. 
As a result, the GPU memory usage and the run-time speed of attention-based one-stage detectors are much larger and slower than SELSA, which makes  them impossible to work in real-world applications.

To overcome this problem, one straightforward solution is to reduce the $N_q$ for one-stage detectors. In two-stage detectors, RPN predicts proposals around the object regions to remove the background regions and thus produce a small number of proposals. However,  RPN is removed in one-stage detectors, leading to a heavy computational cost. Here, we ask a question: can we utilise temporal information in videos to filter out background regions in a frame and reduce  $N_q$? This question leads us to design the location prior network which uses the detection results of the previous frame to find possible foreground regions on the current frame.

\begin{table}[t]
\caption{Runtime dissection of a one-stage detector.}
\label{tab:part}
\centering
\begin{tabular}{ccccc}
\toprule
\multicolumn{2}{c}{Part}        & {\ Specification\ }  & {\ Runtime (ms)} & {\ \ Ratio\%\ \ }  \\
\midrule
\multicolumn{2}{c}{Backbone\ }   & R-50          & 7.7          & 18.0 \\
\multicolumn{2}{c}{Neck}   & FPN           & 0.5          & 1.2  \\
\midrule
\multirow{5}{*}{Head\ \ } & Level 1 & $s$=8            & 27.7  & 64.9 \\
                      & Level 2 & $s$=16           & 3.9    & 9.1 \\
                      & Level 3 & $s$=32           & 1.4    & 3.3  \\
                      & Level 4 & $s$=64           & 0.8    & 1.9  \\
                      & Level 5  & $s$=128          & 0.7   & 1.6 \\
                    \midrule
\multicolumn{2}{c}{All} &- &42.7 & 100.0\\
                      \bottomrule
\end{tabular}
\end{table}

\subsection{Bottleneck 2: Detection Heads on Low Feature Levels}
\label{subsec:bottleneck2}
To further boost the speed and perform an efficient one-stage video object detector, 
we dissect the run time consumption in each part of the one-stage detector. We use FCOS as a representative for the demonstration. As shown in \tab{part}, the backbone and the neck module can run relatively fast. Nearly 80\% of the running time is spent on the detection head. Then, we dissect the runtime consumption in the detection head according to feature levels.
Specifically, around 65\% of the running time is spent on the first feature level in the detection heads.
The reason for this computational bottleneck is that the feature maps in low levels have very high resolutions. Therefore, the decoding process, i.e., generating detections for every location, is very time-consuming. The high-resolution feature maps are demonstrated to be useful for detecting small objects  \cite{fpn,pan,fcos} and they inevitably consume huge computational costs.

However, it is possible to reduce the computational bottleneck on low-level feature maps by utilising the size prior knowledge of videos. Specifically, since the size of objects in consecutive frames changes gradually, 
we can skip the detection head on low-level feature maps for several frames if there is not any small object in the previous frame. Inspired by this observation, we design the size prior network and achieve a very fast one-stage VOD.

\section{Methodology}\label{sec:method}
We introduce two modules, the location prior network (LPN) and the size prior network (SPN), to address the two computational bottlenecks, respectively. In this section, we illustrate the details in the LPN and SPN and FCOS \cite{fcos} as a representative one-stage detector for demonstration.

\begin{figure*}[t]
\begin{center}
\includegraphics[width = 0.9\textwidth]{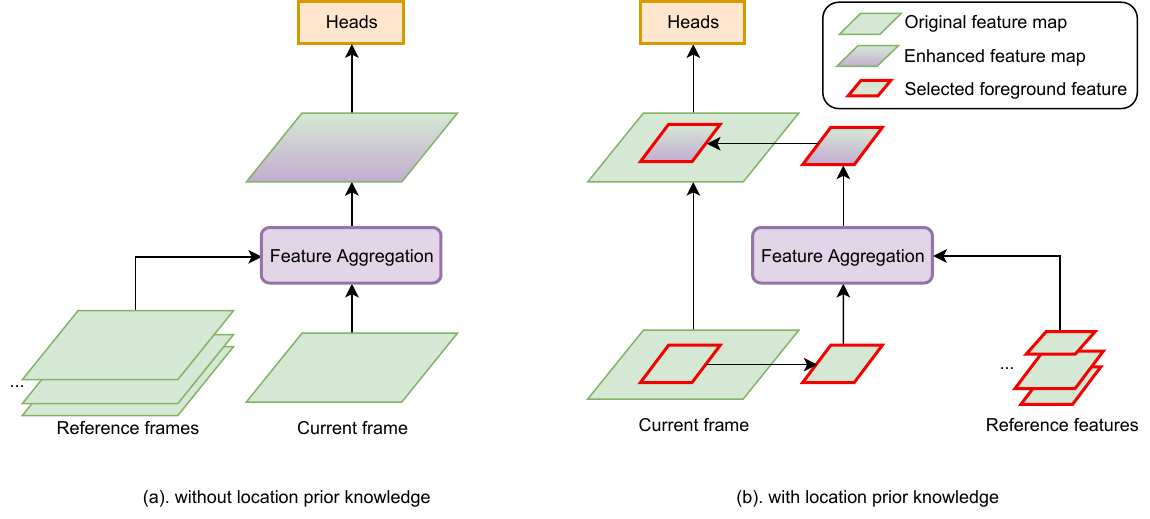}
\end{center}
\caption{(a) shows the process of directly conducting attention-based feature aggregation, where the purple rounded rectangle denotes the attention module. In (a), the input of the attention module is all pixels on the current frame and reference frames. (b) shows the pipeline of using location prior network for feature aggregation, where the red bounding box denotes the propagated bounding boxes from the previous frame. In (b), the input of the attention is foreground pixels on the current frame and the reference frames. \bestviewed}
\label{fig:lpn}
\end{figure*}

\subsection{Location Prior Network}
\label{subsec:lpn}
As analysed in \subsecs{bottleneck1}, reducing $N_q$ is the key of performing an efficient one-stage video object detection with attention-based multi-frame aggregation.
We propose a location prior network (LPN) to select foreground regions in the current frame to conduct partial feature aggregation. The LPN has two steps: First, the foreground region selection guided by the detected bounding boxes from the previous frame; Second, the partial feature aggregation to enhance the selected foreground pixels using attention modules. 

We follow the open-source implementation\footnote[1]{\href{https://github.com/open-mmlab/mmtracking/blob/c250394b8a9ca95dae2ad49efe2d92ae450f605a/configs/vid/selsa/selsa_faster_rcnn_r50_dc5_1x_imagenetvid.py\#L24}{https://github.com/open-mmlab/mmtracking}} of SELSA\cite{selsa} and use 14 random selected reference frames. The comparisons between conducting feature aggregation with and without LPN are shown in \figa{lpn}.

\subsubsection{Foreground Region Selection.}

Given detected bounding boxes of the previous frame, pixels within validated bounding boxes $\{D^v\}$ are regarded as foreground pixels for partial feature aggregation. Here, validated bounding boxes $\{D^v\}$ denote the detected boxes whose classification scores are greater than 0.5. If there is not a validated bounding box, the partial feature aggregation is skipped.
Specifically, we project boxes in $\{D^v\}$ to each feature level by dividing the stride $s$ of the level, e.g., 4 and 8.
Then, we generate a binary mask $M\in \mathbb{R}^{1\times H \times W}$ of foreground regions for every level. The value of the mask at a location $(x,y)$ is assigned as 1, if $(x,y)$ falls into any validated bounding boxes. Otherwise, it is set to 0. 
In addition, before generating $M$, boxes in $\{D^v\}$ are resized with an adjustment ratio $r$ to control the computational overhead.

\subsubsection{Partial Feature Aggregation.}
Given the binary mask $M$, the pixel of location $(x, y)$ is regarded as a foreground pixel if $M(x,y)$ = 1. Then, foreground pixels are enhanced with features from reference frames $\mathbf{F}^r$ via attention modules.
At last, the enhanced pixels are used to replace the pixels of the feature maps $F$ in the same location. Specifically, the enhanced feature maps $\hat{F}$ are computed as follows:

\begin{align}
\begin{aligned}
\hat{F}(x,y) = \begin{cases} \mathbf{A}[F(x, y), \mathbf{F}^r] \quad\quad \text{if }  M(x, y) = 1, \\
                    F(x, y) \quad\quad\quad\quad\quad \text{else, } \end{cases}
\end{aligned}
\end{align}
where $\mathbf{A}(\cdot, \cdot)$ is defined as in \equa{att} and $(x,y)$ enumerates all locations on the feature maps.
Finally, the new feature maps of the current frame are used for detection heads to predict bounding boxes.

\subsubsection{Training and Inference.}
At the training stage, we adopt the strategy of temporal dropout used in  \cite{fgfa} to randomly select two support frames within the same video of the current frame $I_t$. Then the ground truth boxes are used to generate the foreground mask and select query and key pixels. The whole network is optimised with the detection losses computed on the current frame in an end-to-end manner.
During the inference stage, at a time step, the foreground mask of the current frame $I_t$ is propagated from the detection results of the previous frame $I_{t-1}$. The key set consists of the pixels within detected bounding boxes on the reference frames.


\begin{figure*}[t]
\begin{center}
\includegraphics[width = 0.8\textwidth]{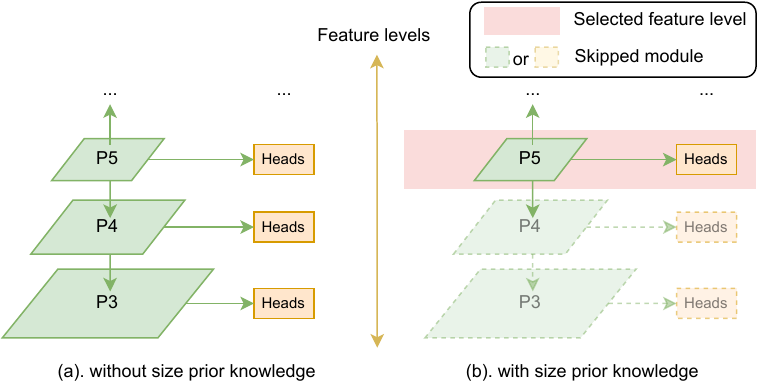}
\end{center}
\caption{(a) shows a normal detection process on multi-level feature maps where all levels of feature maps are passed to detection heads. (b) shows the detection process guided by the size prior network. The pink box denotes the feature level which the bounding boxes of the previous frame are generated from. In the current frame, computations on feature levels not in the pink box are skipped, denoted with the transparent boxes and dotted lines. \bestviewed}
\label{fig:spn}
\end{figure*}

\subsection{Size Prior Network}
The second computational bottleneck for efficient one-stage detection is due to the computations on low-level feature maps.
We introduce the size prior network (SPN) to skip computations on low-level feature maps in unnecessary frames. 
Specifically, after the detection process of a frame at time step $t$, SPN selects validated bounding boxes $\{D^v_t\}$. Here, the validated boxes are obtained in the same way as mentioned in our LPN \subsecs{lpn} (classification score $>0.5$). For the next $T$ time steps, the detection heads are conducted only on the feature levels that boxes in $\{D^v_t\}$ are generated from.

For example, in a video frame $I_t$, the validated bounding boxes $\{D^v_t\}$ are generated from the top feature level, e.g., P5, which indicates there might not be small objects in the following frames and thus the detection process is unnecessary to be conducted on low-level feature maps. For these frames, skipping the huge computations on low-level feature maps does not affect the detection accuracy. In practice, we set the detection frame interval $T$=7 for a good speed-accuracy trade-off. The comparisons between the detection process with and without SPN are shown in \figa{spn}.

\section{Experiments}
\label{sec:exp}
In this section, we first conduct experiments to verify the effect of the proposed location prior network (LPN) and size prior network (SPN). We use three modern one-stage detectors as the base detector, i.e., FCOS \cite{fcos}, CenterNet \cite{centernet}, and YOLOX \cite{yolox}, which also demonstrate the good compatibility of our method. At last, we compare our method with SOTA video object detection methods.

\subsection{Experimental Setting}
\subsubsection{Dataset.} We evaluate our method on the ImageNet VID  \cite{imagenet} dataset  which  contains  3,862  training and  555  validation  videos. We follow previous approaches  \cite{mega,selsa,fgfa,dff} and train our model on the overlapped 30 classes of ImageNet VID and DET set.  Specifically, we sample 15 frames from each video in VID dataset and  at  most  2,000  images  per  class  from  DET  dataset  as our training set. 

\subsubsection{Backbone and Detection Architecture.} For the backbone models, we use ResNet-101 \cite{resnet} in FCOS and CenterNet, and DarkNet-53 \cite{yolox} in YOLOX. 
For the neck architectures, FPN \cite{fpn} and PAN \cite{pan} are used for FCOS and YOLOX, respectively.
For CenterNet, following the same structure as in the original paper, the neck is built with 3 de-convolutional layers with 256, 128, and 64 channels, respectively. One 3×3 deformable convolutional layer is added before each de-convolution with channels 256, 128, and 64, respectively. In FCOS and YOLOX, shared detection heads are performed on all levels of feature maps to generate  the  detections.  Specifically, there  are  two  separate branches inside the detection head.  Each branch contains four 3×3 convolution layers with 256 channels and one 3×3 convolution layer for predicting regression  and classification  result  maps. 
In CenterNet, the detection head is built with a 3×3 convolutional layer with 64 channels followed by a 1×1 convolution with corresponding channels (e.g., number of classes) to
generate detection outputs.

\subsubsection{Training and Inference Details.}
We train all models on 4 Tesla V100 GPUs. 
For FCOS, images are resized to a shorter side of 600 pixels and the longer side less than or equal to 1000 pixels.
With batch size equal to 4, we train the network for 3 epochs using the SGD optimizer (momentum:  0.9, weight decay:  0.0001).  The learning rate is 10$^{-3}$ for the first 2 epochs and 10$^{-4}$ for the last epoch. 
For CenterNet, we follow  \cite{chp} to resize the input images to 512 $\times$ 512. 
Random flip and random scaling from 0.6 to 1.4 are used as data augmentation and SGD is used as the optimizer. We train the network with a batch size of 32 and a learning rate of 10$^{-4}$ for 50 epochs followed by a learning rate of 10$^{-5}$ for 30 epochs.
For YOLOX, we follow  \cite{yolox} to resize the input images to 640 $\times$ 640 and use additional data augmentations, including MixUp \cite{mixup}, Mosaic, RadomCrop, etc. We train the network with batch size 32 using the SGD optimizer. The initial learning rate is set to 10$^{-3}$ with a cosine learning rate schedule for 80 epochs.
In the inference phase, we resize the input image in the same way as in the training phase and reserve the top 100 confident detections per frame.

\subsection{Effect of the Location Prior Network (LPN)}
\label{subsec:lpn_exp}

We first adapt LPN to three modern one-stage detectors to verify its effectiveness. Then, we conduct experiments using different bounding box adjustment ratios $r$ in LPN to find an optimal speed-accuracy trade-off.

\subsubsection{LPN on Various One-stage Detectors.}

\begin{table}[t]
\centering
\caption{Effect of the LPN on different one-stage detectors.}
\label{tab:lpn}
\begin{tabular}{lcccccccc} 
\toprule
Method    & LPN & AP   & AP50  & AP75 & APs  & APm  & APl  & FPS   \\
\midrule
FCOS  \cite{fcos}     &     & 49.8 & 73.3. & 54.6 & 10.1 & 22.4 & 56.1 & 25.1  \\
FCOS  \cite{fcos}      & \checkmark   & 54.1 & 79.8  & 59.5 & 10.5 & 28.3 & 60.1 & 20.4  \\
CenterNet  \cite{centernet} &     & 49.3 & 73.4  & 55.4 & 10.2 & 16.3 & 56.5 & 40.9  \\
CenterNet  \cite{centernet} & \checkmark   & 53.4 & 79.8  & 58.5 & 10.3 & 21.9 & 59.5 & 35.5  \\
YoloX-M  \cite{yolox}  &     & 49.4 & 69.4  & 55.4 & 11.1 & 25.0 & 55.4 & 39.7  \\
YoloX-M  \cite{yolox}  & \checkmark   & 53.3 & 75.1  & 58.1 & 11.6 & 30.2 & 58.9 & 35.8  \\
\bottomrule
\end{tabular}
\end{table}

To study how the proposed location prior network (LPN) influences the overall performance, we integrate LPN on three one-stage detectors and show the results in \tab{lpn}.
LPN can improve the performance of all three detectors. 
Specifically, single-frame FCOS achieves 49.8\% of AP. By utilising LPN to conduct multi-frame feature aggregation, the performance of FCOS+LPN is significantly improved by 4.3\% to 54.1\% of AP.
Similarly, we can observe improvements in experiments of using LPN on more lightweight detectors, i.e., CenterNet and YOLOX-M.
CenterNet+LPN and YOLOX-M+LPN boost their baseline performance from 49.3/49.4\% of AP to 53.4/53.0\% of AP, respectively.
These experimental results demonstrate the effectiveness and compatibility of the LPN.

\begin{table}[t]
\centering
\caption{Effect of the bounding box adjustment ratios $r$ in LPN.}
\label{tab:ratio}
\begin{tabular}{lccccccc} 
\toprule
$r$ & AP   & AP50 & AP75 & APs  & APm  & APl  & FPS   \\
\midrule
0.5   & 53.6 & 78.5 & 57.9 & 9.6  & 27.0 & 58.5 & 21.4  \\
0.8   & 54.1 & 79.8 & 59.5 & 10.5 & 28.3 & 60.1 & 20.4  \\
1.0   & 54.2 & 80.0 & 59.7 & 10.5 & 28.5 & 60.2 & 19.1  \\
1.2   & 54.2 & 80.0 & 59.8 & 10.6 & 28.4 & 60.2 & 17.5  \\
1.5   & 54.2 & 79.9 & 59.9 & 10.9 & 28.4 & 60.2 & 14.7  \\
\bottomrule
\end{tabular}
\end{table}

\subsubsection{Effect of the Box Adjustment Ratios $r$.} We show the experimental results on FCOS to study the effect of employing different bounding box adjustment ratios $r$. A smaller $r$ results in fewer selected foreground pixels to be enhanced and thus leads to a faster speed. In contrast, a larger $r$ selects more pixels to be enhanced but causes a slower speed. In our experiments, we vary $r$ from $0.5$ to $1.5$.
Our LPN achieves the optimal speed-accuracy trade-off when $r$ = 0.8, i.e., 54.1\% AP and 20.4 FPS. Once the adjustment ratio is larger than 0.8, the performance is less affected by the change in the adjustment ratios, but the run time keeps increasing. In our experiments, we set $r$ as 0.8 by default.

\subsection{Effect of the Size Prior Network (SPN)}
\label{subsec:spn_exp}

Similar to the experimental design in \subsecs{lpn_exp}, we conduct experiments by adding the size prior network (SPN) to two modern one-stage detectors to verify its effectiveness and compatibility. We use FCOS \cite{fcos} and YOLOX \cite{yolox} as representatives because they work with multi-level feature maps. Then, we conduct experiments using different frame intervals in SPN to find an optimal setting.

\subsubsection{SPN on various One-stage Detectors.}

\begin{table}[t]
\centering
\caption{Effect of the SPN on different one-stage detectors.}
\label{tab:spn}
\begin{tabular}{lccccccccc} 
\toprule
Method  & LPN & SPN & AP   & AP50 & AP75 & APs  & APm  & APl  & FPS   \\
\midrule
FCOS   & \checkmark   &     & 54.1 & 79.8 & 59.5 & 10.5 & 28.3 & 60.1 & 20.4  \\
FCOS     & \checkmark   & \checkmark   & 53.8 & 76.9 & 58.9 & 9.8  & 27.3 & 59.5 & 26.9  \\
YoloX-S & \checkmark   &     & 53.3 & 75.1 & 58.1 & 11.6 & 30.2 & 58.9 & 35.8  \\
YoloX-S & \checkmark   & \checkmark   & 52.7 & 74.5 & 56.7 & 11.2 & 28.9 & 57.7 & 50.5  \\
\bottomrule
\end{tabular}
\end{table}

The location prior network (LPN) is designed to improve the accuracy of one-stage video object detection by conducting efficient multi-frame feature aggregation. While the size prior network (SPN) mainly focuses on improving the run-time speed by skipping the unnecessary computations in specific feature levels.
Specifically, by introducing SPN, the run-time speed of FCOS+LPN and YOLOX+LPN are improved from 20.4/35.8 FPS to 26.9/50.5 FPS, respectively.
At the same time, the accuracy is still at a comparable level with 53.8/52.7\% of AP.
As illustrated in \subsecs{bottleneck2}, most computations of one-stage detectors happen on feature maps of low levels. Therefore, the run time saved by using SPN is mainly because of the computations saved in video frames where no small objects appear. To further understand the speed improvement, we list the portion of skipped frames and the portion of frames according to different object sizes in the supplementary material.

\subsubsection{Effect of the Temporal Frame Interval $T$.}
To explore the effect of temporal frame interval $T$ in SPN, we show the performance of introducing SPN on FCOS+LPN  under various $T$ settings from 0 to 28 in \tab{interval}. During inference, we first conduct a full detection on all feature levels of the current frame and then we use SPN to skip computations on some feature levels for $T$ following frames.
In the extreme case of $T=0$, full detections are conducted on all frames.
With $T=7$, full detections happen in every 8 frames and partial detections happen in the 7 interval frames. In this setting, the run-time speed is significantly improved from 25.1 FPS to 40 FPS with the performance slightly decreased by 0.2\% to 53.9\% AP.
By increasing the temporal interval $T$ to larger numbers, the run-time speed continuously increases, however, the performance also degrades at the same time.
In practice, we set the temporal interval $T=7$ to obtain a good speed-accuracy trade-off.

\begin{table}[t]
\centering
\caption{Effect of temporal frame interval $T$ in SPN.}
\label{tab:interval}
\begin{tabular}{lccccccc} 
\toprule
$T$ & AP   & AP50 & AP75 & APs  & APm  & APl  & FPS   \\
\midrule
0        & 54.1 & 79.8 & 59.5 & 10.5 & 28.3 & 60.1 & 20.4  \\
7        & 53.8 & 76.9 & 58.9 & 9.8  & 27.3 & 59.5 & 26.9  \\
14       & 53.0 & 75.4 & 56.6 & 9.6  & 26.9 & 58.7 & 28.4  \\
21       & 51.9 & 73.8 & 55.7 & 9.2  & 25.4 & 56.9 & 29.0  \\
28       & 48.5 & 73.3 & 54.0 & 8.9  & 24.6 & 55.4 & 29.5  \\
\bottomrule
\end{tabular}
\end{table}

\subsection{Comparisons with SOTA Methods}

\begin{table}[t]
\centering
\caption{Comparisons with SOTA video object detection methods.}
\label{tab:sota}
\resizebox{\textwidth}{!}{%
\begin{tabular}{lccccccccc}
\toprule
Method         & Base Detector & AP   & AP50 & AP75 & APs  & APm  & APl  & FPS  & Device    \\
\midrule
RDN  \cite{rdn}     & FasterRCNN    & -     & 81.8 & -     & -     & -     & -     & 10.6 & V100      \\
SELSA  \cite{selsa}   & FasterRCNN    &-      & 80.3 &-      & -     & -     &-      & -    & -         \\
LRTR  \cite{lrtr}    & FasterRCNN    & -     & 81.0 &-      & -     & -     &  -    & 10   & Titan Xp  \\
MEGA  \cite{mega}    & FasterRCNN    & -     & 82.9 & -     & -     &  -    &  -    & 8.7  & 2080ti    \\
TFB \cite{tfblender} & FasterRCNN   & -     & 83.8   & -     & -     & -     &-      & 4.9  & 2080ti    \\
MAMBA \cite{mamba}  & FasterRCNN   & -     & \textbf{84.6}   &-      & -     &  -    & -     & 9.1  & Titan RTX    \\
TransVOD \cite{transvod} & Deform. DETR   & -     & 81.9   & -     & -     & -     & -     & -  & -    \\
CHP  \cite{chp}     & CenterNet     & -     & 76.7 & -     & -     & -     & -     & 37   & -         \\
\midrule
\midrule
FasterRCNN*  \cite{fasterrcnn}  &-  & 49.7 & 75.6 & 55.9 & 7.4  & 23.7 & 56.0 & 22.5 & V100      \\
FCOS*  \cite{fcos}  &-  & 49.8 & 73.3 & 54.6 & 10.1 & 22.4 & 56.1 &  25.1    & V100      \\
CenterNet*  \cite{centernet}  &-  & 49.3 & 73.4 & 55.4 & 10.2 & 16.3 & 56.5 &  40.9    & V100      \\
YoloX-M*  \cite{yolox}    &-  & 49.4 & 69.4 & 55.4 & 11.1 & 25.0 & 55.4 &   39.7   & V100      \\
\midrule
RDN*  \cite{rdn}    & FasterRCNN    & 53.4 & 81.2 & \textbf{60.1} & 8.5  & 27.4 & 59.6 & 7.1  & V100      \\
SELSA*  \cite{selsa} & FasterRCNN    & 52.6 & 81.6 & 57.9 & 9.3  & 28.6 & 58.4 & 6.4  & V100      \\
MEGA*  \cite{mega}  & FasterRCNN    & 53.2 & 82.4 & 59.2 & 9.1  & 29.4 & 59.1 & 5.3  & V100      \\
\midrule
Ours(+LPN)    & FCOS      & \textbf{54.1} & 79.8 & 59.5 & 10.5 & 28.3 & \textbf{60.1} & 20.4 & V100      \\
Ours(+LPN)    & CenterNet & 53.4 & 79.8 & 58.5 & 10.3 & 21.9 & 59.5 & 35.5 & V100      \\
Ours(+LPN)    & YOLOX-M     & 53.3 & 75.1 & 58.1 & \textbf{11.6} & \textbf{30.2} & 58.9 & 35.8 & V100     \\
Ours(+LPN+SPN) & YOLOX-M  & 52.7 & 74.5 & 56.7 & 11.2 & 28.9 & 57.7 & \textbf{50.5} & V100     \\
\bottomrule
\end{tabular}
}
\end{table}

We compare our method with SOTA VOD methods, and the results are shown in \tab{sota}.
As most SOTA methods are neither report with the run-time speed nor test on the same device, we re-implement recent SOTA methods and test them on our device for fair comparisons. The methods with $*$  denote our re-implementation versions.
In addition, most SOTA methods are based on the two-stage  detector, FasterRCNN, while we propose an efficient one-stage VOD method. 

All results are reported with the same backbone R-101 except the YOLOX whose backbone is DarkNet-53.
Overall, our method achieves a better speed-accuracy trade-off. In particular, our method with FCOS achieves 54.1\% of AP at 20.4 FPS, which makes 0.7\% accuracy improvement and nearly $3\times$ speed improvement over the best competitor RDN. 
As expected, our method can achieve very good efficiency. Compared with the only existing one-stage VOD method, CHP, our method with CenterNet makes a 3.1\% improvement of AP50. Considering the run-time speed, we adapt LPN and SPN on the YOLOX-M detector. Our method, YOLOX-M+LPN+SPN, runs very fast at 50.5 FPS on V100 GPU, much faster than other existing VOD methods. Moreover, the YOLOX-M+LPN+SPN model achieves good performance with 52.7\% of AP, which is comparable to the SOTA method SELSA in our implementation. 
These results highlight the advantages of our method in terms of accuracy and speed.
It is worth noting that, in our implementation, the two-stage FasterRCNN and the one-stage FCOS, CenterNet, and YOLOX achieve 49.7/49.8/49.3/49.4\% of AP, respectively. The comparable performance of these base detectors demonstrates that the superior performance of our method is solely gained from the modules we proposed instead of the replacement of base detectors.

\section{Conclusion}
By comprehensive analysis, we indicate the computational cost is the underlying obstacle to achieving efficient one-stage video object detection.
To address the computational bottlenecks, we propose a simple yet effective framework that can  be  incorporated into  various one-stage detectors. Specifically, our method consists of two novel modules: (1) The location prior network that selects the foreground regions of the current frame using the detection results of the previous frame; (2) The size prior network to skip unnecessary computations on the low-level feature maps when there is not any small object appears.
Extensive experiments are conducted, and excellent experimental results demonstrate the superior effectiveness, efficiency, and compatibility of our method.

%
%
\bibliographystyle{splncs04}
\bibliography{egbib}
\end{document}

%% file: macro.tex
\usepackage{color} 
\usepackage[normalem]{ulem}
\usepackage{multirow}
\usepackage{graphicx}
\usepackage{color}

\def\figa#1{Fig. \ref{fig:#1}}

\def\equa#1{Eq. (\ref{equ:#1})}

\def\tab#1{Table \ref{tab:#1}}

\def\sect#1{Section \ref{sec:#1}}

\def\subsecs#1{\S\ref{subsec:#1}} 

\def\bestviewed{\textit{Best viewed in colour.}}

